# Confidence Regularized Masked Language Modeling using Text Length[1]

**Seunghyun Ji and Soowon Lee[2]**
Soongsil University
soryhyun@omofictions.com, swlee@ssu.ac.kr

## Abstract

Masked language modeling is a widely used method for learning language representations, where the model predicts a randomly masked word in each input. However, this approach typically considers only a single correct answer during training, ignoring the variety of plausible alternatives that humans might choose. This issue becomes more pronounced when the input text is short, as the possible word distribution tends to have higher entropy, potentially causing the model to become overconfident in its predictions. To mitigate this, we propose a novel confidence regularizer that adaptively adjusts the regularization strength based on the input length. Experiments on the GLUE and SQuAD benchmarks show that our method improves both accuracy and expected calibration error.

## 1 Introduction

As learning distributed language representations before training improves the performance of natural language processing models (Turian et al., 2010), learning language representation - so-called pre-training - became an essential training procedure. Especially when learning bidirectional contextualized representations is recommended, such as natural language understanding, masked language modeling (MLM, Devlin et al., 2019) is primarily chosen for learning language representations.

Masking a random token in the input and recovering the original text is the main process of MLM. This can make the model learn contextualized representations effectively. While calculating loss in MLM, words in the whole vocabulary except a single word become incorrect answers. Some guess is made that this may cause inefficiency in training (Clark et al., 2020). Also, although different words can fill the masked position (Zhou et al., 2019), MLM ignores these predictable words and calculates the loss with a single answer.

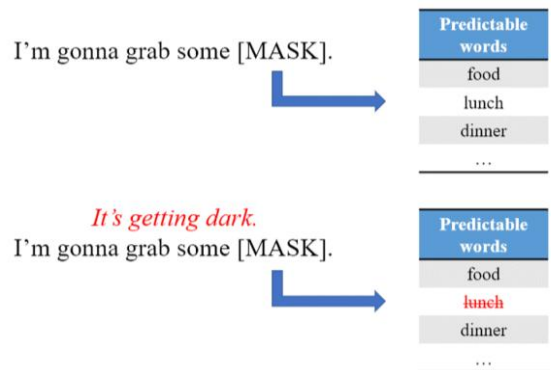

Figure 1: Two examples of entropy of probability distributions in a masked position. In (a), various words such as 'food', 'lunch', and 'dinner' can fill the masked position. Conversely, in (b), when the sentence is extended by adding the text 'It's getting dark.', the likelihood of the word 'lunch' from (a) becomes negligible. Although the answers in (a) and (b) may be the same, the distributions of predictable words differ significantly.

Confidence regularizers such as label smoothing (Szegedy et al., 2016) and confidence penalty (Pereyra et al., 2017), can be employed to prevent the model from being overconfident. These methods improve performance in various tasks, such as machine translation (Vaswani et al., 2017), and can also serve as calibration methods (Müller et al., 2019). However, whether applying a confidence regularizer to MLM improves the target task performance is not yet studied. Besides, when the existing confidence regularizers are applied to representation learning, such as image classification tasks, its representations become less transferable (Kornblith et al., 2021).

---

[1] This paper was submitted in ACL 2023, also had been submitted as a graduation thesis on Dec 2022. Reference: https://www.riss.kr/link?id=T16600089

[2] Corresponding author.

In this paper, we propose a novel confidence regularizer that penalizes the confident output when the input text is short. This is based on our intuition that, since a short text contains less context information than a long text, a distribution of words that can fill in the masked position would be high when the input text is short. Label smoothing can be the alternative. However, since label smoothing forces a specific output on noncorrect labels, the concern of inefficiency caused by massive vocabulary can be intensified.

In experiments with GLUE dataset and SQuAD dataset, we found that our method improved the task performance over MLM and an existing confidence penalty regularizer (Pereyra et al., 2017). We also verified that our method significantly decreased the expected calibration error (ECE, Naeini et al., 2015), a widely known metric that shows how well a model is calibrated.

## 2 Related Works

### 2.1 Confidence Regularization in Representation Learning

Label smoothing (Szegedy et al., 2016) is one of the confidence regularizers that improve the performance of various tasks. Suppose the cross entropy between the model prediction $\hat{y}$ and the answer $y$ as $H(y, \hat{y})$, the label smoothed loss $\mathcal{L}_{LS}$ can be described as:

$$\mathcal{L}_{LS} = (1-\alpha)H(y, \hat{y}) + \alpha H(u, \hat{y}) \quad (1)$$

where $u$ is a uniform distribution with support of $\hat{y}$ and $\alpha \in [0,1]$ is a hyperparameter.

As $H(u, \hat{y}) = D_{KL}(u||p) + H(u)$ where $D_{KL}$ is KL divergence and $H(u)$ is the entropy of $u$, the dynamics of label smoothing forces the model to output a certain value across the whole vocabulary. On the other hand, the confidence penalty (Pereyra et al., 2017) regularized loss $\mathcal{L}_{CP}$, without forcing the model to output a certain value across the whole vocabulary, can be described as:

$$\mathcal{L}_{CP} = H(y, \hat{y}) - \beta H(\hat{y}) + C \quad (2)$$

where $C$ is a constant with respect to model parameters and $\beta$ is a hyperparameter (Meister et al., 2020).

While confidence regularizers enhance the performance of various tasks, they have been criticized for causing a model to learn less transferable features, which leads to poor performance in target tasks (Kornblith et al., 2021; Müller et al., 2019). However, when label noise is present in the dataset, label smoothing has been shown to learn much more transferable knowledge than vanilla cross-entropy (Lukasik et al., 2020).

### 2.2 Applicability of Calibration Methods

The importance of model calibration has been mentioned when the decision should be made with a certain confidence. If the model assigns the same probability as an expert's inference, the model can make a decision that should be made with low risk (Jiang et al., 2012). Confidence regularizers can help achieve this objective (Müller et al., 2019).

Recently, calibration methods have been used not only for model reliability. When confidence calibration is adopted in adversarial training, the accuracy of a model can be improved (Stutz et al., 2020). Massive models such as GPT-3 (Brown et al., 2020) can also be calibrated to be unbiased, which leads to performance improvements (Zhao et al., 2021).

The correlation between the model calibration error and other metrics, such as performance or robustness, has not been proven. However, controlling the strength of a confidence regularizer by adversarial robustness, which is the metric that estimates how easily input data can be attacked, makes the model more calibrated (Qin et al., 2021). Moreover, comparing BERT and RoBERTa (Liu et al., 2019), which are trained with similar algorithms and different datasets, the model with better performance showed more calibrated results (Desai and Durrett, 2020).

## 3 Confidence Penalty using Text Length

The problem with MLM is that, although there may be more than one word that can be in a masked position, those probabilities are ignored and make the model overconfident in the single answer. Although confidence regularizers can give uncertainty to the answer, giving uncertainty in all cases may induce less transferable features. Instead, when a noise is in the label, applying a confidence regularizer would help the model to learn transferable representations (Lukasik et al., 2020).

Since various words can fill in the masked position, a single word is different from the true distribution that people would think. Especially when the input text contains little information, which may be proportional to the text length, the entropy of the 'predictable' word distribution is high. When the information is added by lengthening the text, there fewer words can fill the masked position. A simple example can express this in Figure 1.

In summary, when the input text is short, there is a large discrepancy between a label and human reasoning. This can be considered as 'noise' since its concept is different from the correct concept (Angluin and Laird., 1988). In this sense, we propose a novel method that strengthens the regularizer when the input text is short. We did not determine the target entropy directly since the amount of information provided may vary from person to person.

In detail, while in MLM, a random word would be chosen and masked in the tokenized text $x$ and the masked word becomes the hard label answer $y$. Then, the vanilla cross entropy is used for calculating loss with $y$ and the inferred distribution $\hat{y}$. Instead, our method trains the model with the proposing loss $\mathcal{L}_{CP-L}$. As regularizing entropy from beginning of the training would disturb learning, we added a hinge loss to the confidence penalty (Pereyra et al., 2017). Then, for the entropy threshold, our method uses the length of the tokenized text, $len(x)$, divided by the model's maximum input token length, $maxlen$.

$$\mathcal{L}_{CP-L} = H(y, \hat{y}) + \max(0, \beta(1 - r) - H(\hat{y})) \quad (3)$$

$$r = len(x)/maxlen \quad (4)$$

where $\beta$ is an hyperparameter.

If $len(x) = maxlen$, no penalty is given even if the answer is overconfident during training. On the other hand, if the length of input text is short and the training has progressed to some extent, a penalty is applied depending on the confidence of $\hat{y}$. By controlling the hyperparameter $\beta$, it is possible to determine the magnitude and speed at which the penalty is given. Since neural network models perform operations in batches, we do not calculate the text length individually but use the max-pooled length.

| Methods (steps) | GLUE Average | SQuAD 1.1 (F1) | SQuAD 2.0 (F1) |
|---|---|---|---|
| MLM-50k | 76.01 | 78.46 | 62.52 |
| MLM-150k | 78.29 | 82.45 | 64.68 |
| MLM-250k | 79.37 | 83.92 | 66.32 |
| CP-AvgL-50k | 75.98 | 79.01 | 63.24 |
| CP-AvgL-150k | 77.96 | 82.35 | 65.05 |
| CP-AvgL-250k | 79.20 | 83.83 | 67.06 |
| CP-L-50k | 75.89 | 79.57 | 62.80 |
| CP-L-150k | 78.66 | 81.65 | 64.85 |
| CP-L-250k | **79.53** | **84.16** | **67.08** |

Table 1: The comparison between our method and existing methods with GLUE and SQuAD 1.1/2.0 datasets. We fine-tuned 7 models with fixed random seeds for each task and verified the average score on validation sets. Our method shows ~0.7 points improvement over MLM and traditional confidence penalty (CP-AvgL).

## 4 Experiments

### 4.1 Experimental Settings

**Pre-training** Following Devlin et al. (2019), we selected the English BookCorpus (800M words after WordPiece tokenization) (Zhu et al., 2015) and the English Wikipedia [3] as a pre-training corpora. Since our method utilizes text length, we didn't concatenate sentences in a document to get various short sentences. Then, we made a batch by grouping data with a similar length of tokenized text. The open-source function '*group_by_length*' by Huggingface [4] was used to implement this feature. This led to an overhead in GPU computations, but in about ~2 epochs, we achieved the average score of GLUE benchmark (Wang et al., 2019) as much as Devlin et al. (2019). We adopted the AdamW optimizer (Loshchilov and Hutter, 2019), with a learning rate of 2e-4, and trained the model for up to 250k steps with a maximum token length of 512. The pre-training procedure was conducted on 16 A100 GPUs.

**Fine-tuning** We evaluated methods on the GLUE benchmark (Wang et al., 2019) and SQuAD 1.1/2.0 datasets (Rajpurkar et al., 2016, 2018). Following Devlin et al. (2019), we excluded WNLI from tasks of GLUE benchmark. We reported Matthew's correlation score for CoLA, Pearson correlations for STS-b, F1 score for SQuAD 1.1/2.0, and accuracy scores for the other tasks. Due to the nature of BERT, we changed some seeds and re-trained the model in cases where it failed to learn.

---

[3] https://huggingface.co/datasets/wikipedia

[4] https://github.com/huggingface/transformers

| Methods | [10, 50) | [50, 200) | [200, 512] |
|---|---|---|---|
| MLM | 2.17 | 2.08 | 1.27 |
| CP-AvgL | 2.17 | 1.28 | 1.11 |
| CP-L, ours | **1.79** | **1.02** | **1.04** |

Table 2: Expected calibration error (ECE) on GLUE, SQuAD 1.1/2.0 dataset while MLM with several length intervals of the input text. Our method showed the lowest ECE on every interval, which denotes most calibrated.

| Methods | GLUE Average | SQuAD 1.1 (F1) | SQuAD 2.0 (F1) |
|---|---|---|---|
| MLM | 64.80 | 67.38 | 53.19 |
| LS-L | **66.23** | 65.38 | 50.39 |
| CP-L, ours | 65.27 | **67.55** | 53.18 |

Table 3: The comparison of fine-tuned models with average GLUE score and SQuAD 1.1/2.0 scores. All models are fine-tuned with BERT-mini. Label smoothing using text length (LS-L) showed better performance in GLUE, but severely poorer performance in SQuAD 1.1/2.0.

### 4.2 Fine-tuning Results

To compare pre-training methods, including our method, we pre-trained the BERT-base using MLM, our mtehod (CP-L), and the traditional confidence penalty (CP-AvgL). Note that CP-AvgL is the same as CP-L except that the *average token length of a dataset* is used for $len(x)$. To find the best hyperparameter for our method, we pre-trained BERT-mini (Turc et al., 2019) and selected the setting which made the best result for fine-tuning tasks. As a result, we set hyperparameter $\beta$ to 2.

Table 1 shows the results when BERT-base is pre-trained with each method and then fine-tuned with GLUE and SQuAD datasets. We reported the average score of 7 different models trained with fixed random seeds. CP-L showed the best performance compared with MLM and CP-AvgL when pre-trained up to 250k steps. Considering that CP-AvgL has the same regularizing strength as CP-L, our method is better for learning contextualized representations than the traditional confidence penalty.

### 4.3 Expected Calibration Error Results

To verify whether model is calibrated or not, the expected calibration error (ECE, Naeini et al., 2015), which is widely known metric, can be used. Using a $m$-th confidence interval $B_m$, ECE is described as:

$$ECE = \sum_{m=1}^{M} \frac{|B_m|}{n} |acc(B_m) - conf(B_m)| \quad (5)$$

where $n$ is total sample number, $acc()$ is the accuracy and $conf()$ is the confidence of the model.

We sampled 1,000 texts each by 3 intervals of text length from GLUE and SQuAD 1.1/2.0 datasets. Then we calculated the ECE when models which are pre-trained until 250k steps by performing MLM. We averaged ECE of 7 trials with different fixed random seeds. Table 2 shows that our method showed the lowest ECE in every text length interval, which denotes the most calibrated model. Especially our method showed a better calibration score than CP-AvgL, which was set to have the same regularizing strength ($\beta = 2$).

### 4.4 Label Smoothing using Text Length

Although label smoothing can intensify the problem of MLM, we implemented label smoothing using text length (LS-L) and ran experiments to verify this. As regularizing strength of label smoothing is determined by hyperparameter $\alpha$ we control the strength using text length ratio $r$:

$$\alpha = T(1-r)^2 \quad (6)$$

where $T$ is a hyperparameter. When the token length approaches 0, $\alpha$ increases up to $T$ and naturally increases the smoothing strength. We squared the text length ratio term for similar controlling intensity with our method (CP-L), which regularizes the confidence in a log scale. Pre-training BERT-mini and choosing the best hyperparameter setting, the hyperparameter $T$ was set to 0.05.

We pre-trained BERT-mini up to 150k steps and compared the results of fine-tuned models. Table 3 shows that LS-L achieved the best performance in GLUE benchmark but severely poorer performance in SQuAD 1.1/2.0, which primarily consists of long text. To sum up these results, LS-L induced the model to learn less transferable representations when the input text is long.

## 5 Conclusion

In this paper, we proposed a novel confidence regularizer for language representation learning to solve the problem that models learn with an overconfident label while in MLM. Instead of regularizing confident output in all cases, our method gives a penalty when the short text is given and the output distribution is confident. With experiments, we verified that our method makes the model learn more transferable representations and makes the model to be more calibrated than traditional methods. Due to the simplicity of our method, various combinations and research would remain unconstrained.

## 6 Limitations

This research has two limitations. At first, our method cannot consider some exceptional cases when the text length is not related to the entropy of distributions. For example, predictable words in "*I [MASK] gonna grab a lunch*" would be small; therefore, they formulate low entropy. However, our method does not take this into account and forces the model to be underconfident. To deal with this issue, various rule-based methods can be used. However, as our approach aims to build a regularizer with a simple formula, which would not increase computations, we skipped applying those methods.

Secondly, our experiment was run in small epochs with BERT-base, therefore incurs limited expressivity. This is due to the varying text token length, which would require the flexible batch size to resolve the overhead in GPU. We cannot assure that our method beats MLM in large parameter settings (e.g., BERT-large) or other models (e.g., XLNet (Yang et al., 2019), ELECTRA (Clark et al., 2019), TACO (Fu et al., 2022) as our experiments are limited to BERT-base. However, since confidence regularizers also improve massive models (Szegedy et al., 2016, Vaswani et al., 2017), our method can be efficacious in massive models. Besides, calibration methods also improve the performance of various tasks if the model's parameter is massive, like GPT-3 (Zhao et al., 2021).

## Ethics Statement

We respect the ACL Code of Ethics and comply with the ACL Ethics Policy while in research.

## A Pre-training Details

**Datasets** We used the English BookCorpus (800M words after WordPiece tokenization) (Zhu et al., 2015) and English Wikipedia as pre-training corpus. BookCorpus is a dataset split with a sentence, and Wikipedia is split into a document. Because using the raw Wikipedia dataset would cause various problems, we re-split the Wikipedia dataset with a paragraph. As the maximum token length is 512, paragraphs over 512 were truncated, and no multi-paragraph data was used. We expected this would cause no problem because excluding *<sep>* token while in MLM improved the performance of finetuned models (Liu et al., 2019).

**Hyperparameters** Hyperparameter settings of BERT-base and BERT-mini are listed in Table 4.

## B Fine-tuning Details

**Datasets** The datasets we used for fine-tuning includes GLUE benchmark (Wang et al., 2019) and SQuAD 1.1/2.0 (Rajpurkar et al., 2016, 2018). Those include the Corpus of Linguistic Acceptability (CoLA) (Warstadt et al., 2018), Question Natural Language Inference (QNLI) (Rajpurkar et al., 2016), Recognizing Textual Entailment (RTE) (Bentivogli et al., 2009), Quora Question Pairs (QQP [5]), Multi-Genre Natural Language Inference (MNLI) (Williams et al., 2018), the Stanford Sentiment Treebank (SST) (Socher et al., 2013), Microsoft Research Paraphrase Corpus (MRPC) (Dolan and Brockett, 2005), and the Semantic Textual Similarity Benchmark (STS-B) (Cer et al., 2017). Number of examples of these are listed in Table 5.

**Hyperparameters** Hyperparameter settings for fine-tuning BERT-base and BERT-mini are listed in Table 5.

## C Full Results

**Fine-tuning Results of BERT-base** GLUE Fine-tuning results of BERT-base are described in Table 6.

**Fine-tuning Results of BERT-mini** Fine-tuning results of BERT-mini are described in Table 7. In results of various hyperparameter settings of PE-T, ($\beta = 2.5$) showed the highest average score. However, as ($\beta = 2$) hyperparameter setting showed higher score in SQuAD 1.1/2.0, we selected ($\beta = 2$) setting to safely apply to BERT-base experiments. In experiments of LS-T, all hyperparameter settings showed severely low score in SQuAD 1.1/2.0 compared to MLM.

| Hyperparameters | BERT-base | BERT-mini |
|---|---|---|
| Hidden size | 768 | 256 |
| Number of layers | 12 | 4 |
| Number of attention heads | 12 | 4 |
| Adam $\epsilon$ | 1e-8 | 1e-8 |
| Adam $\beta_1$ | 0.9 | 0.9 |
| Adam $\beta_2$ | 0.999 | 0.999 |
| Dropout probability | 0.1 | 0.1 |
| Layer norm $\epsilon$ | 1e-12 | 1e-12 |
| Total training steps | 250k | 150k |
| Warmup steps | 2.5k | 1.5k |
| Peak learning rate | 2e-4 | 5e-4 |
| Batch sizes | 512 | 576 |
| Seed number | 42 | 42 |

Table 4: Hyperparameter settings of BERT-base and BERT-mini.

---

[5] https://quoradata.quora.com/First-Quora-Dataset-Release-Question-Pairs

| Datasets | Number of examples (train/validation/test) | Hyperparameter settings of BERT-base/BERT-mini | | |
|---|---|---|---|---|
| | | Epochs | Learning rate | Batch size |
| CoLA | 8551/1043/1063 | | | |
| MNLI | 392702/9815/9796 | | | |
| MRPC | 3668/408/1725 | 6/6 | 5e-5/1e-4 | |
| QNLI | 104743/5463/5463 | | | 32/32 |
| QQP | 363846/40430/390965 | | | |
| RTE | 2490/277/3000 | 12/12 | 1e-5/2e-5 | |
| SST2 | 67349/872/1821 | 6/6 | 5e-5/1e-4 | |
| STS-b | 5749/1500/1379 | 12/12 | 2.5e-5/5e-5 | |
| SQuAD 1.1 | 87599/10570/- | 3/5 | 3e-5/3e-4 | 32/48 |
| SQuAD 2.0 | 130319/11873/- | 3/5 | 3e-5/3e-4 | 32/48 |

Table 5: Number of examples of datasets and hyperparameter settings for fine-tuning BERT-base and BERT-mini.

| Methods (steps) | CoLA | MNLI | MRPC | QNLI | QQP | RTE | SST2 | STSb | GLUE Average |
|---|---|---|---|---|---|---|---|---|---|
| MLM-50k | 47.50 | 77.92 | 79.52 | 84.99 | 89.60 | 57.09 | 88.61 | 82.86 | 76.01 |
| MLM-150k | 55.59 | 79.98 | 80.74 | 86.58 | 90.01 | 58.90 | 90.43 | 84.15 | 78.29 |
| MLM-250k | 58.01 | 80.70 | 83.75 | 87.07 | 90.23 | 59.36 | 90.63 | 85.24 | 79.37 |
| CP-L-50k | 46.82 | 78.14 | 79.48 | 84.82 | 89.55 | 55.80 | 89.01 | 83.53 | 75.89 |
| CP-L-150k | 55.79 | 80.04 | 83.05 | 86.34 | 90.07 | 58.84 | 90.27 | 84.89 | 78.66 |
| CP-L-250k | 57.71 | <u>80.90</u> | <u>84.07</u> | 87.00 | 90.28 | <u>59.93</u> | 90.71 | <u>85.66</u> | <u>79.53</u> |
| CP-AvgL-50k | 45.30 | 77.83 | 77.42 | 85.18 | 89.67 | 60.65 | 88.68 | 83.10 | 75.98 |
| CP-AvgL-150k | 54.80 | 79.44 | 80.18 | 86.35 | 89.99 | 57.50 | 90.71 | 84.77 | 77.96 |
| CP-AvgL-250k | <u>59.13</u> | 80.77 | 82.46 | <u>87.14</u> | 90.26 | 57.92 | <u>90.91</u> | 85.00 | 79.20 |

Table 6: GLUE Fine-tuning results of BERT-base.

| Methods (Hyperparameters) | GLUE Average | SQuAD 1.1 | SQuAD 2.0 |
|---|---|---|---|
| MLM | 64.80 | 67.38 | 53.19 |
| CP-L ($\beta = 1.0$) | 64.86 | 67.34 | 53.34 |
| CP-L ($\beta = 2.0$) | 65.27 | <u>67.55</u> | 53.18 |
| CP-L ($\beta = 2.5$) | 65.93 | 67.45 | 52.96 |
| CP-L ($\beta = 3.0$) | 65.70 | 66.89 | <u>54.02</u> |
| CP-L ($\beta = 4.0$) | 65.22 | 66.81 | 53.03 |
| LS-T ($T = 0.05$) | <u>66.23</u> | 65.38 | 50.39 |
| LS-T ($T = 0.1$) | 66.01 | 62.64 | 51.92 |
| LS-T ($T = 0.2$) | 64.91 | 60.02 | 48.48 |
| LS-T ($T = 0.3$) | 64.83 | 62.20 | 50.64 |

Table 7: Fine-tuning results of BERT-mini.